\def\year{2021}\relax
\documentclass[letterpaper]{article} 
\usepackage{aaai21}  
\usepackage{times}  
\usepackage{helvet} 
\usepackage{courier}  
\usepackage[hyphens]{url}  
\usepackage{graphicx} 
\usepackage{natbib}  
\usepackage{caption} 
\usepackage{multirow}
\usepackage{booktabs}

\usepackage{amsfonts,amssymb}
\usepackage{amsmath,bm}

\usepackage[switch]{lineno}

\providecommand{\tabularnewline}{\\}

\title{Learning Comprehensive Motion Representation for Action Recognition}
\author {
    
        Mingyu Wu\textsuperscript{\rm 1\thanks{The first two authors contribute equally.}},
        Boyuan Jiang\textsuperscript{\rm 3\footnotemark[1]},
        Donghao Luo\textsuperscript{\rm 3},
        Junchi Yan\textsuperscript{\rm 1,2\thanks{Correspondence to: Junchi Yan (yanjunchi@sjtu.edu.cn).}},
        Yabiao Wang\textsuperscript{\rm 3}\\
        Ying Tai\textsuperscript{\rm 3},
        Chengjie Wang\textsuperscript{\rm 3},
        Jilin Li\textsuperscript{\rm 3},
        Feiyue Huang\textsuperscript{\rm 3},
        Xiaokang Yang\textsuperscript{\rm 1}\\
}
\affiliations {
    \textsuperscript{\rm 1} MoE Key Lab of Artificial Intelligence, AI Institute, Shanghai Jiao Tong University \\
    \textsuperscript{\rm 2} Department of Computer Science and Engineering, Shanghai Jiao Tong University \\
    \textsuperscript{\rm 3} Youtu Lab, Tencent\\
}
\begin{document}
\maketitle

\begin{abstract}
For action recognition learning, 2D CNN-based methods are efficient but may yield redundant features due to applying the same 2D convolution kernel to each frame. Recent efforts attempt to capture motion information by establishing inter-frame connections while still suffering the limited temporal receptive field or high latency. Moreover, the feature enhancement is often only performed by channel or space dimension in action recognition. To address these issues, we first devise a Channel-wise Motion Enhancement (CME) module to adaptively emphasize the channels related to dynamic information with a channel-wise gate vector. The channel gates generated by CME incorporate the information from all the other frames in the video. We further propose a Spatial-wise Motion Enhancement (SME) module to focus on the regions with the critical target in motion, according to the point-to-point similarity between adjacent feature maps. The intuition is that the change of background is typically slower than the motion area. Both CME and SME have clear physical meaning in capturing action clues. By integrating the two modules into the off-the-shelf 2D network, we finally obtain a Comprehensive Motion Representation (CMR) learning method for action recognition, which achieves competitive performance on  Something-Something V1 \& V2 and Kinetics-400. On the temporal reasoning datasets Something-Something V1 and V2, our method outperforms the current state-of-the-art by 2.3\% and 1.9\% when using 16 frames as input, respectively.
\end{abstract}

\section{Introduction}

Action recognition has been an essential building block for video understanding, whereby the key is to extract powerful spatial-temporal features that contain rich motion information. 3D CNN-based methods~\cite{tran2015learning, carreira2017quo, hara2018can} establish spatial and temporal correlation among video frames by employing 3D convolution kernels. Due to the large number of parameters, these methods suffer from heavy computation and memory costs. To be light-weighted and fast, 2D CNN-based methods~\cite{wang2016temporal,qiu2017learning,lin2019tsm} are proposed.

\begin{figure}[tb!]
    \centering
    \includegraphics[width=.45\textwidth]{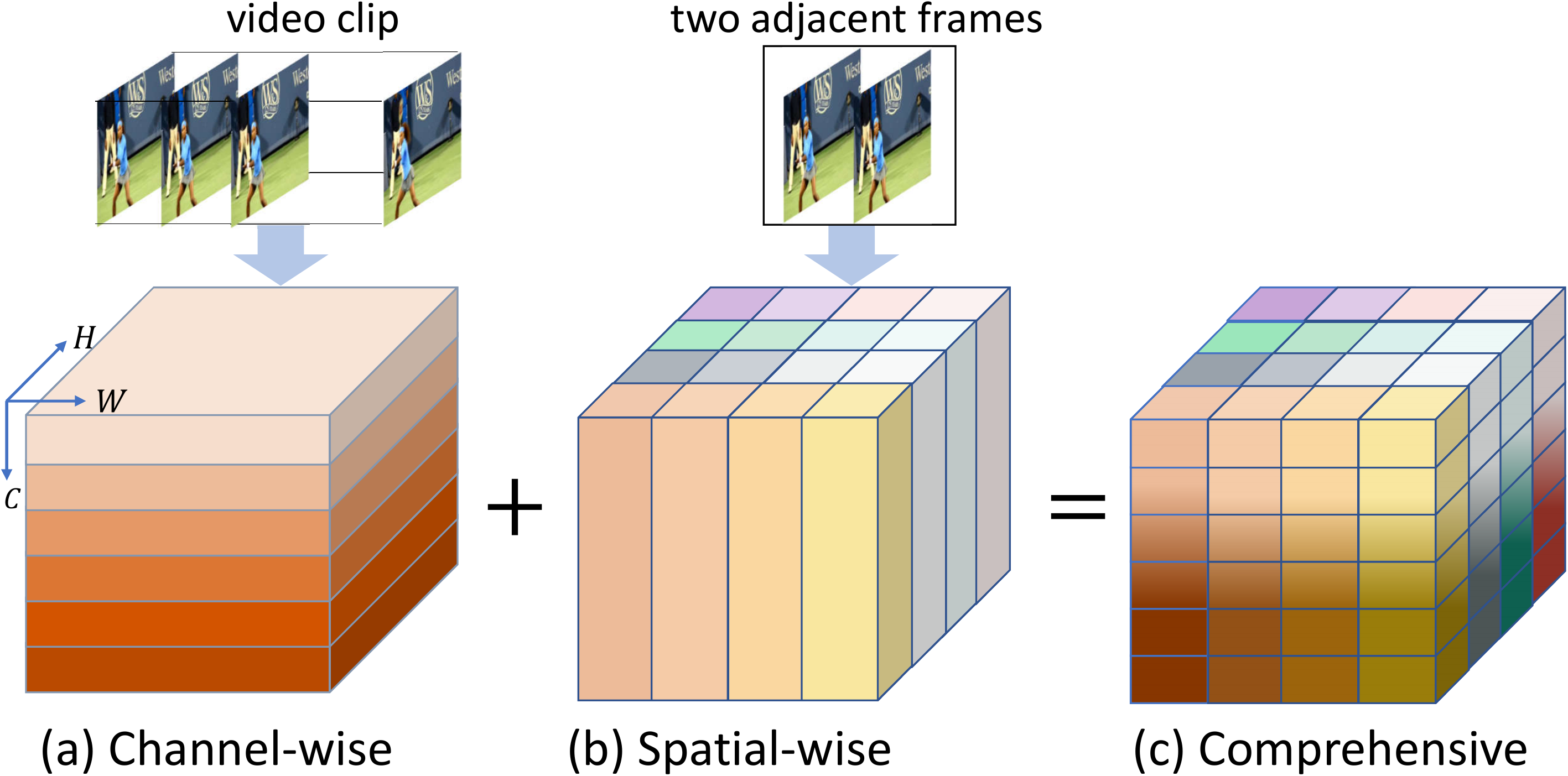}
    \caption{Comprehensive motion representation learning in our method. a) Channel-wise: collect motion information with global temporal receptive field for channel enhancement. b) Spatial-wise: Point-to-point similarity is introduced to describe the motion region between adjacent frames. c) Comprehensive learning: By combining the two aspects together, the features can be enhanced in a fine-grained way.}
    \label{fig:Intro}
\end{figure}

Most 2D CNN-based methods are dedicated to temporal relation modeling, which employs well designed convolutional layers or blocks to process features along the temporal dimension. Since the convolution kernels are shared when processing each frame, the features of each frame may contain excessive redundant background information, leading to covering the crucial motion features. Therefore, motion enhancement for 2D CNN-based methods has attracted much attention in recent years. It focuses on utilizing the inter-frame information, especially motion clues, to guide the feature generation progress. The goal of these methods is to make the model automatically emphasize what is important for video recognition. Some representative work \cite{wang2016hierarchical,meng2019interpretable,liu2020teinet, li2020tea, weng2020temporal} has proved its effectiveness with great improvement. However, there are still several issues in this direction that deserve rethinking. First, most methods focus on designing channel enhancement module but suffer from the limited temporal receptive field. They only collect information from part of frames to construct a channel enhancement vector, so that some irrelevant motion information is easily mistaken for the vital factor of video to be analyzed. It might be much better to decide which channel is important if we look through the whole action progress. Furthermore, previous methods are intended to perform enhancement only from channel-wise aspect. Features from the same channel are assigned with the same coefficient. It will fail to distinguish the crucial moving objects on the spatial dimension. A similar issue appears if we only enhance features from the spatial dimension, which ignores the fact that different channels play different roles in capturing specific semantic information. Neither the channel-wise or spatial-wise enhancement can provide a fine-grained description for the motion clues individually. We are motivated by the above analysis to propose a Comprehensive Motion Representation (CMR) learning framework. The CMR framework contains two specific designed modules to capture motion information in features: the Channel-wise Motion Enhancement (CME) module and the Spatial-wise Motion Enhancement (SME) module. By fully taking advantage of both channel-wise and spatial-wise enhancement, our model achieves the fine-grained motion enhancement as described in Fig. \ref{fig:Intro}.

Specifically, the CME module is employed to highlight the motion-related channels utilizing the self-gating mechanism~\cite{hu2018squeeze}. Different from previous works, we design a global temporal information fusion mechanism to form the channel gate for each frame. The channel gate is generated by the weighted summation of spatial-compressed feature vectors over every frame. Features that are significantly different from the current frame will be assigned with higher summation weights. Besides, the mechanism is implemented in parallel, thus the motion clues represented by the diversity can be captured quickly and precisely by the channel gate to guide the enhancement process.

Meanwhile, the SME module aims at exploring and strengthening the motion-related region in each frame. By establishing the point-to-point similarity map between adjacent feature maps, we can distinguish the dynamic area from the static background, as the variations can lead to the lower similarity between the corresponding regions. Higher weights are then assigned to the dynamic contents to make the feature better represent the motion information.

We evaluate our method on three benchmark datasets: Something-Something V1 \& V2~\cite{goyal2017something} and Kinetics400~\cite{kay2017kinetics}. Specifically, in temporal reasoning datasets Something-Something V1 and V2, the proposed method vastly outperforms the state-of-the-art by $2.3\%$ and $1.9\%$, respectively, which indicates the effectiveness of CME and SME modules for video action recognition. Besides, in the scene reasoning dataset Kinetics-400, our method also achieves competitive results. The contributions of our work can be summarized:

1) We propose a Channel-wise Motion Enhancement (CME) module to adaptively enhance motion-related channels with a global temporal receptive field. Compared with the previous works that search for motion clues within a limited temporal range, our method achieves an effective trade-off between speed and performance.

2) We propose a Spatial-wise Motion Enhancement (SME) module to explicitly highlight the motion area in the feature maps according to the point-to-point similarity between two adjacent feature maps. It is proven to be simple but effective than the previous work on utilizing the inter-frame information to represent the motion information.

3) Both CME and SME are designed with clear physical meaning, and they complement each other's strengths. We combine CME and SME to learn comprehensive motion representation and our approach achieves state-of-the-art performance on three benchmark datasets.

\section{Related Work}

\subsubsection{Video Action Recognition.}
Compared to image recognition, video action recognition requires to utilize the temporal information along with spatial one. Therefore, a simple method is to utilize 3D convolution kernel to extract spatial and temporal features jointly \cite{tran2015learning, carreira2017quo, hara2018can}. Although 3D ConvNets show effectiveness in spatial-temporal modeling and achieve state-of-the-art performance, they suffer from heavy computing and training difficulty. 

To address the limitation of 3D ConvNets, a series of 2D CNN-based methods have been proposed recently for efficient video action recognition~\cite{lin2019tsm, qiu2017learning,lin2019tsm, tran2018closer, liu2020teinet,jiang2019stm}. Among them, P3D~\cite{qiu2017learning} and R(2+1)D~\cite{tran2018closer} decompose the 3D convolution kernel into a 2D convolution for spatial information extracting and a 1D convolution for temporal modeling. TSM~\cite{lin2019tsm} further replaces the 1D convolution with a non-parameter channel shift module to capture the temporal relationship. However, these 2D-based methods only take how to model the temporal evolution of frames, neglecting the influence of redundant information between frames. Therefore, in this paper, we focus on learning comprehensive motion enhanced representation which contains less redundant information, rather than temporal relation modeling.

\subsubsection{Attention and Gating Mechanism.}
Attention mechanism has been widely used in image recognition~\cite{wang2017residual, woo2018cbam, hu2018squeeze}. CBAM \cite{woo2018cbam} proposes the combined use of spatial and channel attention for image recognition. SENet \cite{hu2018squeeze} devises a sub-branch to adaptively recalibrate channel-wise features by explicitly modeling inter-dependencies between channels. For action recognition, attention mechanism is also adopted for discriminative feature learning \cite{diba2018spatio,liu2020teinet, weng2020temporal,li2020tea}. STC~\cite{diba2018spatio} simply employs the fully connected layers to generate the channel and spatial attention for 3D ConvNets, it suffers from the problems that the enhancing branches are all designed at the channel level and each frame shares the same attention weights. TEINet \cite{liu2020teinet} and TEA \cite{li2020tea} use the difference of two adjacent feature maps to generate the channel-wise modulation weights. However, only the limited temporal receptive field is considered in TEINet and TEA to emphasize the motion information in the video, which limits the performance. TDRL \cite{weng2020temporal} proposes a Progressive Enhancement Module (PEM) to extend the temporal receptive field of enhancement learning and achieves encouraging performance. However, in PEM, the calculation of the channel recalibrate factor is inefficient. At each time step, it relies on the state of the previous time step, leading to serial computing, and the temporal receptive field of the first few frames is still limited. We propose two modules to efficiently suppress redundant information and enhance motion information from the channel and spatial dimension separately, achieving better recognition performance.

\section{Proposed Method}
In this section, we first introduce the Channel-wise Motion Enhancement module, as used to adaptively enhance important channels and suppress redundant ones from the global temporal receptive field. Then we present the Spatial-wise Motion Enhancement module to highlight features in the motion area. Finally, we describe how to integrate two modules into off-the-shelf 2D CNN-based methods to learn comprehensive motion enhanced representation for actions.

\subsection{Channel-wise Motion Enhancement Module}
\begin{figure}[tb!]
    \centering
    \includegraphics[width=0.46\textwidth,height=0.3\textheight]{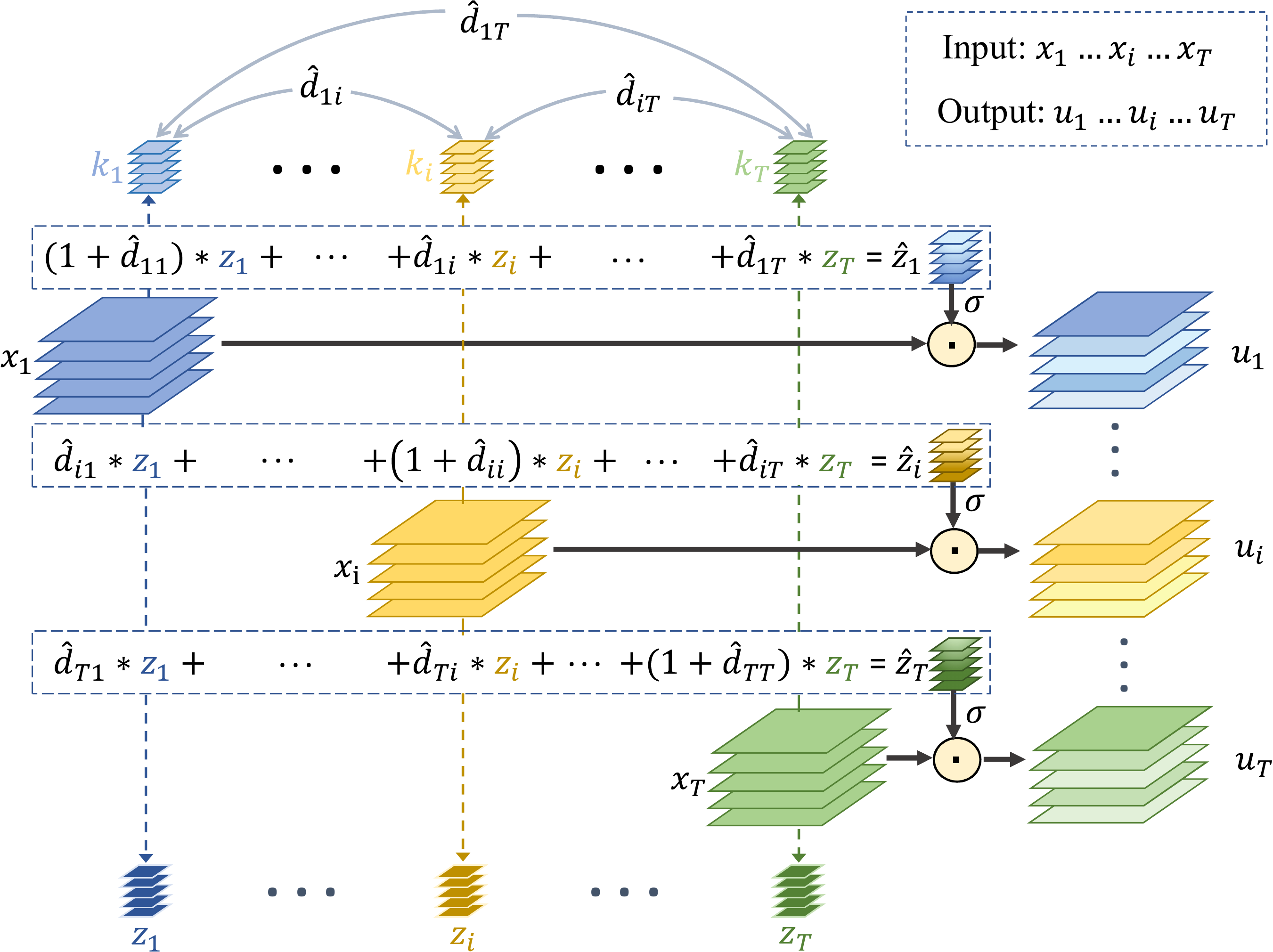}
    \caption{Illustration of CME module. Features of each frame $\mathbf{x}_i$ is first mapped to an intermediate representation vector $\mathbf{z}_i$ through GAP and $1 \times 1$ convolution. Then we accumulate information from other frames to current frame based on discrepancy measurement to form the new intermediate representation vector $\mathbf{\hat{z}}_i$ , \textit{i.e.}, $\mathbf{\hat{z}}_i=\mathbf{z}_i+\sum_{j=1}^T \hat{d}_{ij}\mathbf{z}_j$. Finally, a sigmoid function $\sigma(\cdot)$ with another $1 \times 1$ convolution is applied to convert $\mathbf{\hat{z}}_i$ to channel modulation weights $\mathbf{a}_i$.}
    \label{fig:CME}
\end{figure}

It is known that different channels of feature map tend to describe distinct semantic patterns. Some channels focus on encoding background patterns, while others focus on foreground or motion patterns. For video action recognition, motion-related patterns are more important than static background patterns. To better capture pivotal patterns, we propose the Channel-wise Motion Enhancement Module (CME), to enhance the discriminative channels and suppress the useless ones with the self-gating mechanism. As shown in Fig. \ref{fig:CME}, the input of CME is the feature maps $\{\mathbf{x}_t\}_{t=1}^T$ from the preceding layer where $\mathbf{x}_t \in \mathbb{R}^{C \times H \times W}$ denotes the feature map of frame $t$ and the enhancement operation is:
\begin{equation}
    \mathbf{u}_t = \mathbf{x}_t \odot \mathbf{a}_t, 1 \leq t \leq T,
\end{equation}
where $\mathbf{a}_t \in \mathbb{R}^C$ is the channel enhancement vector, $\odot$ is the channel-wise multiplication operation, and $\mathbf{u}_t$ is the enhanced feature map of $t$-th frame. To obtain channel enhancement vector $\mathbf{a}_t$, one straightforward method is to apply the scheme proposed by SENet \cite{hu2018squeeze}. However, SENet is designated for image recognition tasks, and when generating a channel enhancement vector, it processes each frame of videos independently without considering the information from other frames. Some previous works \cite{liu2020teinet,li2020tea,weng2020temporal} attempt to utilize inter-frame differences to determine which channel should be enhanced. However, only the limited temporal receptive field is considered in these methods, which may ignore those motion patterns that can only be observed from multiple frames. Therefore, to better capture motion information, one should take information from \textit{all the other frames} into account when generating the current frame's channel enhancement vector. 

The process of generating enhancement vector $\mathbf{a}_t$ can be divided into three steps. The first step is to yield channel descriptor $\mathbf{z}_t$ for each frame individually. Since our goal is to select motion-sensitive channels and detailed spatial information is not crucial here, we first apply global average pooling over the spatial dimension of $\mathbf{x}_t$ to obtain channel-wise statistics $\mathbf{\overline{x}}_t$. Then a $1 \times 1$ convolution is employed to capture the channel-wise relationship:
\begin{equation}
    \mathbf{z}_t = {\rm Conv}(\mathbf{\overline{x}}_t, \mathbf{W}_1), 1 \leq t \leq T,
\end{equation}
where $\mathbf{W}_1$ is the learnable parameters of the convolution for feature transformation and the dimension of the output channel is $\frac{C}{r_1}$. $\mathbf{z}_t$ is the intermediate representation used to produce the channel enhancement vector. However $\mathbf{z}_t$ only contains information from current frame $t$. 

The second step is to merge information from other frames' intermediate representation into the current frame's intermediate representation to help determine which channel should be enhanced. The fusion coefficient of two frames is calculated based on the minus of dot-product similarity. To better calculate the similarity, we first utilize a $1 \times 1$ convolution with parameters $\mathbf{W}_2$ for feature transformation and the channel dimension of $\mathbf{k}_t$ is $\frac{C}{r_2}$:
\begin{equation}
    \mathbf{k}_t = {\rm Conv}(\mathbf{\overline{x}}_t, \mathbf{W}_2), 1 \leq t \leq T.
\end{equation}
Based on $\{\mathbf{k}_t\}_{t=1}^T$, we can calculate the discrepancy between every two frames:
\begin{equation}
    d_{tj} = -\mathbf{k}_t \mathbf{k}_j^\top, 1 \leq t,j \leq T,
\end{equation}
where $d_{tj}$ denotes the discrepancy between frame $t$ and $j$. We then normalize the discrepancy vector using softmax function, \textit{i.e.}, $\sum_{j=1}^T \hat{d}_{tj}=1$ to form fusion coefficients and then perform information fusion:
\begin{equation}
    \mathbf{\hat{z}}_t = \mathbf{z}_t+\sum_{j=1}^T \hat{d}_{tj} \mathbf{z}_j,  1 \leq t \leq T, 
    \label{fusion}
\end{equation}
where $\mathbf{\hat{z}}_t$ is the fused intermediate representation containing both intra-frame and inter-frame information. The intuition behind Eq. \ref{fusion} is that when performing temporal fusion, we expect that the representation of the current frame contains more information from variant frames. Here we adopt a residual scheme to maintain the knowledge of itself.

Finally, another $1 \times 1$ convolution with parameters $\mathbf{W}_3$ is used to recover the channel dimension of $\mathbf{\hat{z}}_t$ to $C$, and the channel enhancement vector can be obtained by using the sigmoid activation $\sigma$:
\begin{equation}
    \mathbf{a}_t = \sigma({\rm Conv}(\mathbf{\hat{z}}_t, \mathbf{W}_3)), 1 \leq t \leq T.
\end{equation}

The CME can be efficiently implemented with matrix operations and we discuss how the values of $r_1$ and $r_2$ affect the performance in our experiments.

\begin{figure}[tb!]
    \centering
     \includegraphics[width=.45\textwidth]{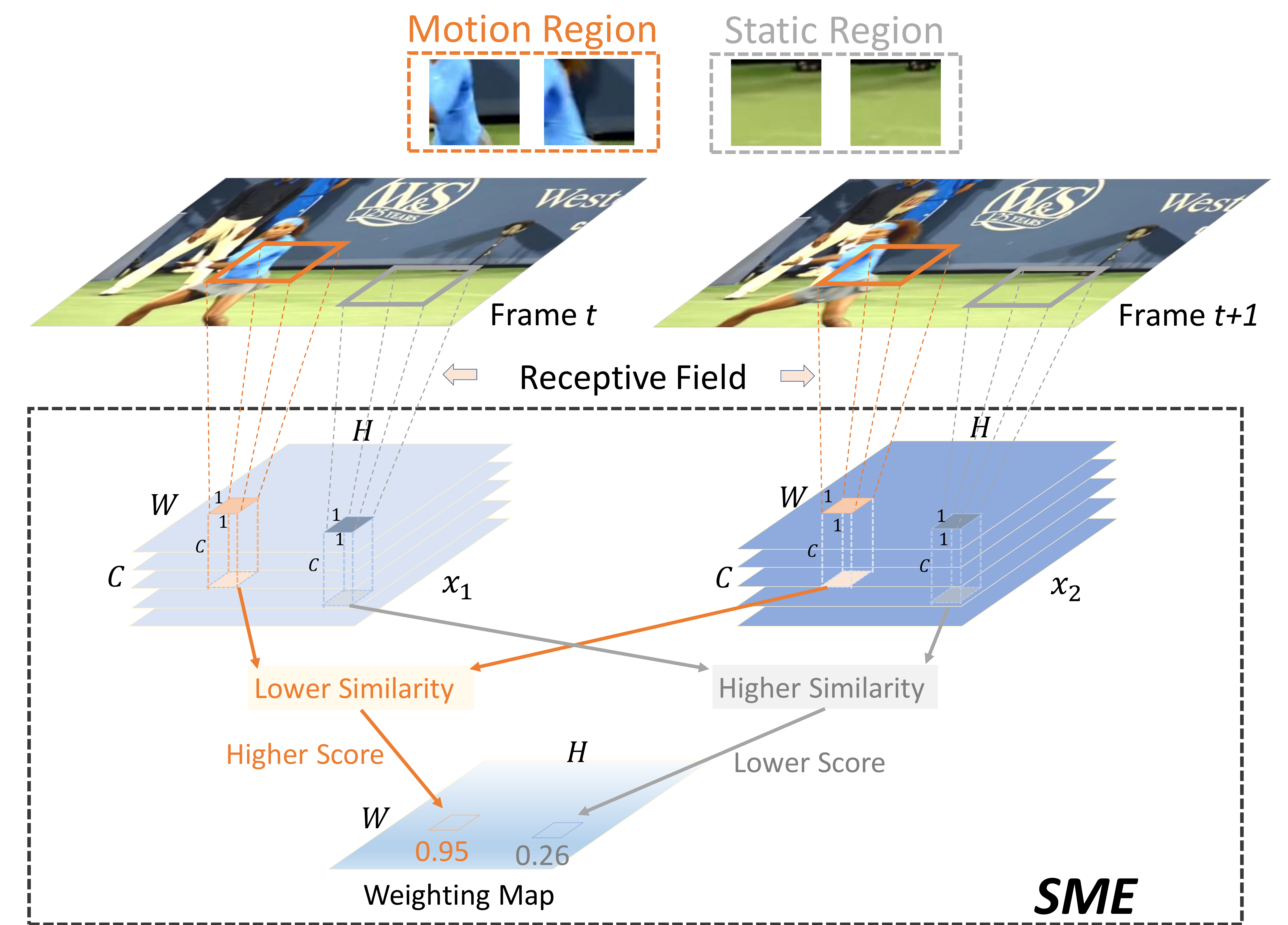}
    \caption{Illustration of the SME module. A spatial weighting map is obtained by calculating the point-to-point similarity between two adjacent feature maps to highlight the features in the motion area.}
    \label{fig:SME}
\end{figure}

\subsection{Spatial-wise Motion Enhancement Module}
CME allows the network to enhance discriminative channels adaptively and suppress redundant ones with a self-gating mechanism at the channel level. However, for the spatial dimension of a specific channel, the modulation weight of each spatial point is the same, which ignores the spatial motion information. Therefore, we devise a Spatial-wise Motion Enhancement Module to explicitly highlight features in the motion area. It serves as the complement of CME.

The input of SME is also a set of frame-level feature maps $\{\mathbf{x}_t\}_{t=1}^{T}$ from the preceding layer. As shown in Fig. \ref{fig:SME}, for a specific feature map $\mathbf{x}_t \in \mathbb{R}^{C \times H \times W}$, each point in $\mathbf{x}_t$ represents a $h \times w$ region at the input frame if the receptive field of current layer is $h \times w$. Therefore, we can calculate the similarity of corresponding points between two adjacent feature maps to represent the motion area of input frames:
\begin{equation}
    \mathbf{s}_t = \eta(\mathbf{x}_t, \mathbf{x}_{t+1}), 1 \leq t \leq T-1,
\end{equation}
where $\eta$ is a function to calculate cosine similarity of corresponding points $(i,j)$ in the feature map between two input feature maps, namely $\eta(\mathbf{x}_t^{(i,j)},\mathbf{x}_{t+1}^{(i,j)}) = \frac{\mathbf{x}_t^{(i,j)} \top \mathbf{x}_{t+1}^{(i,j)}}{\left \| \mathbf{x}_t^{(i,j)}\right \|_2 \left \| \mathbf{x}_{t+1}^{(i,j)}\right \|_2}$. $\mathbf{s}_t$ is the similarity matrix of size $H\times W$. With this matrix, we can explicitly highlight the area where two adjacent feature maps have low similarity. The intuition behind this is that the change of background area is typically slower than the motion area for a video. For a specific spatial point, as all channels share the same attention weight, $\mathbf{s}_t$ will be treated as a matrix that $\mathbf{s}_t \in \mathbb{R}^{C \times H \times W}$ for convenience in subsequent computations. To help convergence, we adopt the following residual learning scheme:
\begin{equation}
    \mathbf{v}_t = {\rm BN}({\rm Conv}(\mathbf{x}_t \circ (1-\mathbf{s}_t)))+\mathbf{x}_t, 1 \leq t \leq T-1, 
    \label{Local}
\end{equation}
where $\mathbf{v}_t$ is the spatial motion enhanced feature. $\circ$ is the element-wise product. Batch normalization and $1 \times 1$ convolution are used to adjust the scale of enhanced feature maps. One may notice that in Eq. \ref{Local}, $t \in [1, T-1]$. To keep the enhanced features' temporal scale consistent with that of the input features, $\mathbf{s}_{T-1}$ is copied to $\mathbf{s}_T$, namely $\mathbf{s}_T = \mathbf{s}_{T-1}$.

\begin{figure}[tb!]
    \centering
    \includegraphics[width=.46\textwidth,height=.21\textheight]{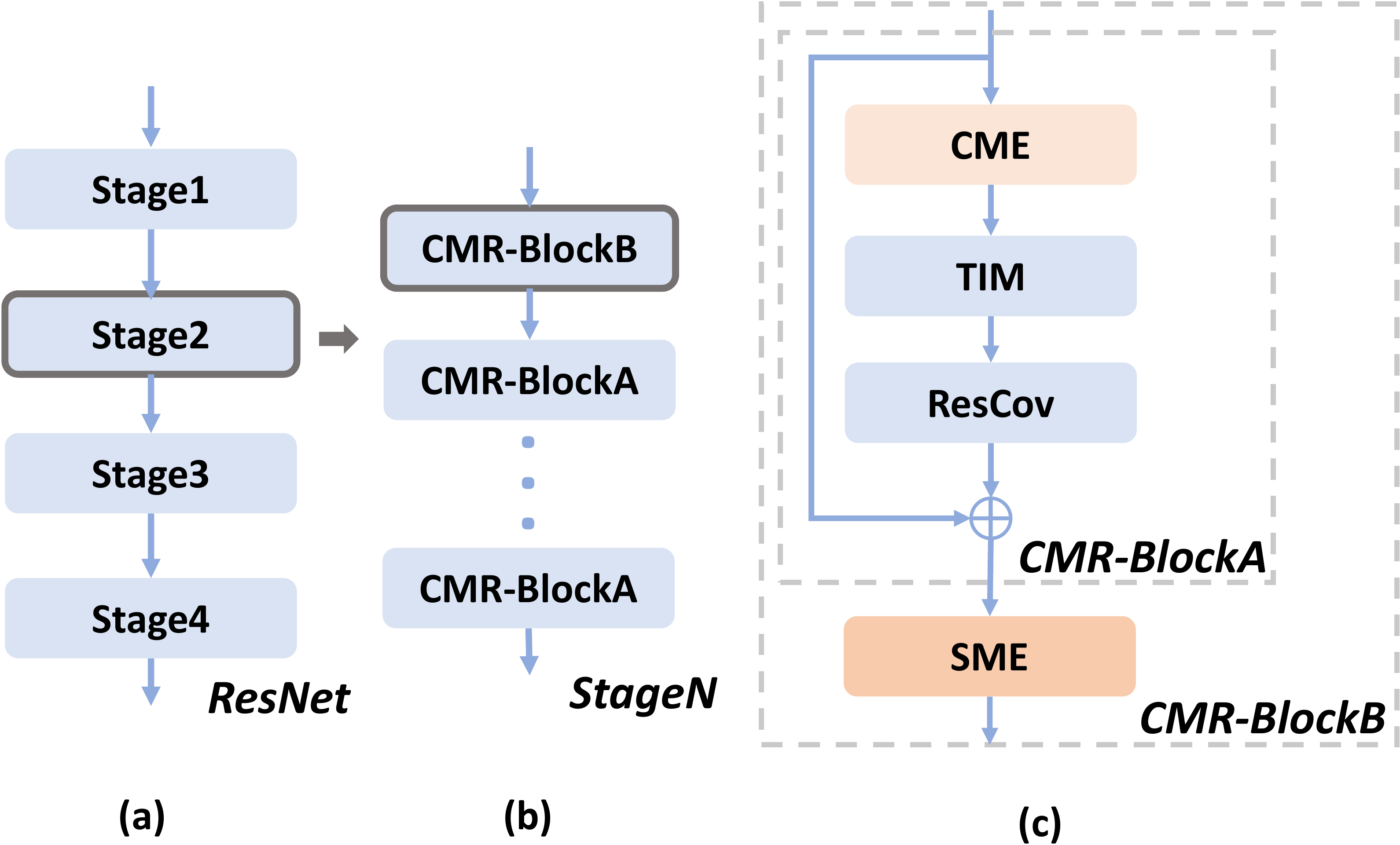}
    \caption{The framework and CMR-Block. a) ResNet: We construct our model on the ResNet structure, which contains four stages. b) StageN: The blocks in each stage are equipped with a combination of TIM, CME, and SME. c) CMR-Block: Two basic blocks are designed for efficiency.}
    \label{fig:pipline}
\end{figure}

\subsection{The Overall Framework}
After introducing the CME and SME module separately, we describe in detail how to integrate them into the network to learn comprehensive motion enhancement features for action recognition. Our proposed modules are efficient, generic, and can be plugged into any existing network to boost the performance. Here we use Resnet \cite{he2016deep} for its popularity in 2D-based action recognition. As shown in Fig. \ref{fig:pipline}, inspired by TEINet \cite{liu2020teinet} and TDRL \cite{weng2020temporal}, we insert the CME module at front of the original ResNet block, to selectively suppress redundant information in the feature maps produced by the preceding layer. Without loss of generality, we also insert the Temporal Interaction Module (TIM) proposed in \cite{liu2020teinet} after CME module for temporal relationship modeling. Following it, we insert operations taken from the original ResNet block, which contains 1$\times$1, 3$\times$3, and 1$\times$1 2D convolutions. We call this modified block as CMR-BlockA. Based on CMR-BlockA, we further insert SME module at the end of the block, as shown in Fig. \ref{fig:pipline} (c), to explicitly enhance motion information at spatial dimension, and we define it as CMR-BlockB. Similar to ResNet architecture, we cascade CMR-Block to form the action recognition network. As shown in Fig. \ref{fig:pipline} (b), in each stage, the first block is replaced with CMR-BlockB, and the rest are with CMR-BlockA. This achieves a trade-off between performance and efficiency as will be shown in the ablation study. 
\begin{table*}[tb!]
    \centering
    \small
\begin{tabular}{llcccccc}
\toprule 
\textbf{Method}& \textbf{Backbone} & \textbf{Frames} & \textbf{Flops} & \textbf{Val$_{1}$} & \textbf{Test$_{1}$} & \textbf{Val$_{2}$} & \textbf{Test$_{2}$}\tabularnewline
\midrule
I3D \cite{wang2018videos} & \multirow{3}{*}{ResNet3D-50} & 32$f$$\times$2 & 306G & 41.6\% & - & - & -\tabularnewline
NL-I3D \cite{wang2018videos} &  & 32$f$$\times$2 & 334G & 44.4\% & - & - & -\tabularnewline 
NL-I3D + GCN \cite{wang2018videos} &  & 32$f\times$ 2 & 606G & 46.1\% & 45.0\% & - & -\tabularnewline
\midrule 
\multirow{2}{*}{ECO \cite{zolfaghari2018eco}} & BNIncep+ & 16$f$ & 64G & 41.6\% & - & - & -\tabularnewline
 & Res3D-18 & 92$f$ & 267G & 46.4\% & 42.3\% & - & -\tabularnewline
\midrule 
\multirow{2}{*}{TSM \cite{lin2019tsm}} & \multirow{2}{*}{ResNet2D-50} & 8$f$ & 33G & 45.6\% & - & 59.1\% & -\tabularnewline
 &  & 16$f$ & 65G & 47.2\% & 46.0\% & 63.4\% & 64.3\%\tabularnewline
\midrule 
\multirow{2}{*}{TEI \cite{liu2020teinet}} & \multirow{2}{*}{ResNet2D-50} & 8$f$ & 33G & 47.4\% & - & 61.3\% & 60.6\%\tabularnewline
 &  & 16$f$ & 66G & 49.9\% & - & 62.1\% & 60.8\%\tabularnewline
\midrule
\multirow{2}{*}{STM \cite{jiang2019stm}} & \multirow{2}{*}{ResNet2D-50} & 8$f$ & 33G & 49.2\% & - & 62.3\% & 61.3\%\tabularnewline
 &  & 16$f$ & 67G & 50.7\% & 43.1\% & 64.2\% & 63.5\%\tabularnewline
\midrule 
\multirow{2}{*}{TEA \cite{li2020tea}} & \multirow{2}{*}{ResNet2D-50} & 8$f$ & 35G & 48.9\% & - & - & -\tabularnewline
 &  & 16$f$ & 70G & 51.9\% & - & - & -\tabularnewline
\midrule
\multirow{2}{*}{GSM \cite{sudhakaran2020gate}} & \multirow{2}{*}{Inception V3} & 8$f$ & 27G & 49.0\% & - & - & -\tabularnewline
 &  & 16$f$ & 54G & 50.6\% & - & - & -\tabularnewline
\midrule
\multirow{4}{*}{TDRL \cite{weng2020temporal}} & \multirow{4}{*}{ResNet2D-50} & 8$f$ & 33G & 49.8\% & 42.7\% & 62.6\% & 61.4\%\tabularnewline
 &  & 16$f$ & 66G & 50.9\% & 44.7\% & 63.8\% & 62.5\%\tabularnewline
 &  & 8$f$$\times$2 & 198G & 50.4\% & - & 63.5\% & -\tabularnewline
 &  & 16$f$$\times$2 & 396G & 52.0\% & - & 65.0\% & -\tabularnewline
\midrule
\multirow{4}{*}{CMR (ours)} & \multirow{4}{*}{ResNet2D-50} & 8$f$ & 33G & \textbf{51.3\%} & \textbf{43.7\%} & \textbf{63.7\%} & \textbf{62.2\%}\tabularnewline
 &  & 16$f$ & 66G & \textbf{53.2\%} & \textbf{47.4\%} & \textbf{65.7\%} & \textbf{64.1\%} \tabularnewline
 &  & 8$f$$\times$2 & 198G & \textbf{51.9\%} & \textbf{44.5\%} & \textbf{64.6\%} & \textbf{63.3\%}\tabularnewline
 &  & 16$f$$\times$2 & 396G & \textbf{54.3\%} & \textbf{48.0\%} & \textbf{66.1\%} & \textbf{64.7\%}\tabularnewline
\bottomrule
\end{tabular}
     \small
    \caption{Comparison with the state-of-the-art on Something-Something V1 $\&$ V2. The subscripts of `Val' and `Test' indicate dataset version and top-1 accuracy is reported.}
    \label{tab:comp_sota_sth}
\end{table*}

\section{Experiments}
We test our method on three large-scale benchmark datasets, i.e., Something-Something V1 $\&$ V2~\cite{goyal2017something} and Kinetics-400~\cite{kay2017kinetics}. Furthermore, hyperparameter in our method is discussed. We also conduct ablation study on the temporal reasoning dataset Something-Something V1 to analyze CME and SME's performance individually and visualize each part's effect. Finally, we give runtime analysis to show the efficiency of our method compared with state-of-the-art methods.

\subsection{Datasets}
\subsubsection{Something-Something V1$\&$V2.}
They are two large-scale video datasets for the interaction between people and objects in daily life. They focus on the temporal correlation of the movement in video clips with little complex background information. The representational ability of models for temporal information can be well verified on these two datasets. V1 includes 108,499 video clips, and V2 includes 220,847 video clips. Both of them contain 174 action categories.

\subsubsection{Kinetics-400.}
It contains 400 human action categories in different scenarios, with at least 400 clips for each category. The average duration of each video is around 10 seconds. Since videos in this dataset mostly contain rich background information with a large inter-class difference, Kinetics-400 focuses on verifying the model's capability on scene classification rather than the effectiveness of temporal modeling.

\subsection{Experimental Setup}
We construct our model and conduct experiments based on the structure of ResNet-50~\cite{he2016deep} pre-trained by ImageNet~\cite{krizhevsky2012imagenet}, considering the trade-off between performance and efficiency.

\subsubsection{Training.} 
We preprocess the training samples with the strategy adopted in~\cite{wang2018non}, which is widely used in previous works. The shorter side of RGB images is resized to 256 and then center cropping and scale-jittering are performed. Each frame will be resized to $224\times 224$ before being fed into the neural network. For the Something-Something V1$\&$V2 datasets, we uniformly sample 8 or 16 frames as a clip from all frames of a video. Since videos in the Kinetics dataset are relatively longer, a dense sampling strategy will be applied to generate the input. We uniformly sample 8 or 16 frames from 64 consecutive frames randomly sampled from the whole video. For the Something-Something dataset, we train the model for 50 epochs, set the initial learning rate to 0.01 and reduce it by a factor of 10 at 30, 40, 45 epochs. For Kinetics-400, our model is trained for 100 epochs. The initial learning rate is set to 0.01 and will be reduced by a factor of 10 at 50, 75, and 90 epochs. Stochastic Gradient Decent (SGD) with momentum 0.9 is utilized as the optimizer, and the batch size is 64 for all three datasets.

\subsubsection{Testing.} 
For fairness, we follow the widely used setting adopted in \cite{wang2018non,liu2020teinet,weng2020temporal}. The shorter side of images will be resized to 256. Center cropping will be performed on each input frame when only 1 clip is sampled. In the experiment that contains multiple clips (e.g. $N$), we take three crops from the left, middle, and right of each frame and then uniformly sample $N$ clips ($8f \times N$ or $16f \times N$) in each video for classification individually. The final prediction is obtained by calculating the average classification score of the $N$ clips. For Kinetics-400, $N=10$.

\subsection{Comparison with State-of-the-Arts}
Evaluation is performed on Something-Something V1$\&$V2 and Kinetics-400 datasets. The results on Something-Something V1$\&$V2 are shown in Table~\ref{tab:comp_sota_sth}. Compared to 2D CNN-based methods, our method outperforms the current state-of-the-art TDRL by $1.5\%$ and $1.1\%$ on V1 and V2 under the setting of $8f$, meanwhile $2.3\%$ and $1.9\%$ under the setting of $16f$. The better performance under the $16f$ setting can be explained by the powerful ability to capture discriminative information such as motion information. We also witness an improvement on Kinetics-400 in Table~\ref{tab:comp_sota_k400}. Although our method's accuracy is only $0.2\%$ and $0.1\%$ higher on the setting of 8f and 16f than the 2D-based state-of-the-art method TDRL, our method is much faster than TDRL since the calculating of channel attention weights in TDRL is serial. One may notice that our method is inferior to the 3D-based state-of-the-art method NL+SlowFast \cite{feichtenhofer2019slowfast}. The reasons may be in two-folds. First, our method focuses on modeling motion information in videos. The improvement in the scene reasoning dataset like the Kinetics-400 is not obvious as it on the temporal reasoning datasets. Second, the backbone of NL+Slowfast is 3D ResNet-101, and it utilizes more input frames, which is much computational heavy than our method.

\begin{table*}[tb!]
    \centering
    \small
    \begin{tabular}{llccc}
    \toprule
    \textbf{Method}&\textbf{Backbone}&\textbf{FLOPs$\times$views}&\textbf{Top}-1&\textbf{Top}-5\\
    \midrule
    I3D$_{64f}$~\cite{carreira2017quo}&Inception V1(I)&$108\times$N/A&72.1\%&90.3\%\\
    NL+I3D$_{32f}$~\cite{wang2018non}&ResNet3D-50(I)&$70.5\times 30$&74.9\%&91.6\%\\
    NL+I3D$_{128f}$~\cite{wang2018non}&ResNet3D-50(I)&$282\times 30$&76.5\%&92.6\%\\
    Slowfast~\cite{feichtenhofer2019slowfast}&ResNet3D-50(S)&$36.1\times 30$&75.6\%&92.1\%\\
    Slowfast~\cite{feichtenhofer2019slowfast}&ResNet3D-101(S)&$106\times 30$&77.9\%&93.2\%\\
    NL+Slowfast~\cite{feichtenhofer2019slowfast}&ResNet3D-101(S)&$234\times 30$&\textbf{79.8\%}&\textbf{93.9\%}\\
    SlowFast+RMS$_{32f}$~\cite{kim2020regularization}&ResNet3D-50(S)&N/A$\times$30&76.3\%&92.5\%\\
    \midrule
    R(2+1)D$_{32f}$~\cite{tran2018closer}&ResNet2D-34(S)&$152\times 10$&72.0\%&90.0\%\\
    S3D-G$_{64f}$~\cite{xie2018rethinking}&Inception V1(I)&$71.4\times 30$&74.7\%&93.4\%\\
    TSM$_{8f}$~\cite{lin2019tsm}&ResNet2D-50(I)&$33\times 30$&74.1\%& - \\
    TSM$_{16f}$~\cite{lin2019tsm}&ResNet2D-50(I)&$65\times 30$&74.7\%&91.4\%\\
    STM$_{16f}$~\cite{jiang2019stm}&ResNet2D-50(I)&$67\times 30$&73.7\%&91.6\%\\
    TEINet$_{8f}$~\cite{liu2020teinet}&ResNet2D-50(I)&$33\times 30$&74.9\%&91.8\%\\
    TEINet$_{16f}$~\cite{liu2020teinet}&ResNet2D-50(I)&$66\times 30$&76.2\%&92.5\%\\
    TEA$_{8f}$~\cite{li2020tea}&ResNet2D-50(I)&$35\times 30$&75.0\%&91.8\%\\
    TEA$_{16f}$~\cite{li2020tea}&ResNet2D-50(I)&$35\times 70$&76.1\%&92.5\%\\
    TDRL$_{8f}$~\cite{weng2020temporal}&ResNet2D-50(I)&$33\times 30$&75.7\%&92.2\%\\
    TDRL$_{16f}$~\cite{weng2020temporal}&ResNet2D-50(I)&$66\times 30$&76.9\%&93.0\%\\
    \midrule
    CMR$_{8f}$ (Ours)&ResNet2D-50(I)&$33\times 30$&\textbf{75.9\%}&\textbf{92.3\%}\\
    CMR$_{16f}$ (Ours)&ResNet2D-50(I)&$66\times 30$&\textbf{77.0\%}&\textbf{93.0\%}\\
    \bottomrule
    \end{tabular}
        \small
    \caption{Comparison with the state-of-the-art on Kinetics-400. The notation `I' in the backbone column indicates that the model is pre-trained with ImageNet. `S' denotes the model is trained from scratch.}
    \label{tab:comp_sota_k400}
\end{table*}

\subsection{Hyperparameter Setting}
A pair of parameters $r_1$ and $r_2$ in the CME module need to be adjusted during the experiment, it reflects the degree that the channel is compressed. To ensure the dimension of the input and output of CME remain unchanged, $r_1$ should be equal to $r_2$. Considering that $r_1$ and $r_2$ must be the divisor of the number of feature channels, we train and test our model on Something-Something V1 with $r_1$ and $r_2$ ranged from $1,4,8,16$. In Table~\ref{tab:hyperparameter}, we notice that the accuracy will be seriously reduced if there is no channel dimension reduction and it will also drop as the downsampling rate $r_1$ exceeds a certain range. A reasonable explanation is that when the values of $r_1$ and $r_2$ are too large, the intermediate representation vector loses key information. However, when $r_1$ and $r_2$ are too small, the motion-independent information contained in the intermediate representation vector will dominate, so that the inter-frame difference caused by motion cannot be effectively captured by CME. We finally adopt $r_1 = r_2 = 8$ by default for all experiments, on which we obtain the best trade-off between complexity and accuracy.

\begin{table}[tb!]
    \centering
    \begin{tabular}{ccc}
        \toprule
         $r_1$\&$r_2$&\textbf{FLOPs}&\textbf{Top-1}\\
         \midrule
         1&55G&47.3\% \\
         4&39G&50.6\%\\
         8&33G&\textbf{51.3\%}\\
         16&32G&50.3\%\\
         \bottomrule
    \end{tabular}
    \caption{Hyperparameter setting analysis. $r_1$ and $r_2$ are the channel scaling factors in CME which are set as the same.}
    \label{tab:hyperparameter}
\end{table}

\subsection{Ablation Study}
We validate our framework on Something-Something V1 under the setting of 8$f$. In this section, we report the results using the testing scheme of center crop and one clip.

The temporal interaction module (TIM)~\cite{liu2020teinet} is employed as baseline in our framework for fair comparison with other enhanced methods~\cite{liu2020teinet,li2020tea,weng2020temporal}. It must be noted that our enhanced module can be combined with any temporal modeling method. Another reason that we choose TIM to model the temporal information is that it is proven to effectively boost the accuracy when combining with our CME and SME. As shown in Table~\ref{tab:TIM-baseline}, the absence of TIM will hurt the performance, but our method still improve by 25.4\% over pure ResNet50 even without TIM.

\begin{table}[tb!]
    \centering
    \begin{tabular}{lc}
        \toprule
         \textbf{Model}&\textbf{Top-1}\\
         \midrule
         Res50&18.1\% \\
         Res50$+$CME\&SME&43.5\%\\
         Res50$+$TIM~\cite{liu2020teinet}&46.1\%\\
         Res50$+$TIM$+$CME\&SME&\textbf{51.3\%}\\
         \bottomrule
    \end{tabular}
    \caption{Baseline design analysis. Based on the result, ResNet-50 with TIM is chosen as the baseline for further enhancement by our techniques.}
    \label{tab:TIM-baseline}
\end{table}

\subsubsection{Number of CME and SME Inserted into Network.}
We first expect to figure out how many CME and SME blocks inserted into the network can obtain the best trade-off performance, although both CME and SME are all lightweight. We separately attempt to insert CME and SME at every block of a ResNet stage or only in the first block. From Table \ref{tab:ablation_num} we can see, apart from only inserting CME and SME at the first block of a stage, other settings achieve almost the same top-1 accuracy. Hence we choose to insert CME at every block of a ResNet stage and SME at the first block of a stage (the second row in Table \ref{tab:ablation_num}), as shown in Fig. \ref{fig:pipline} (b), for the best trade-off performance.

\subsubsection{Performance of CME and SME.}
We then conduct a separate study to show the effectiveness of CME and SME in learning motion-related features. The number of CME and SME inserted is based on the trade-off performance. As shown in  Table~\ref{tab:ablation_tim}, the proposed CME and SME achieve almost the same improvement compared with the baseline, which demonstrates that both CME and SME can help to learn motion-related features. Furthermore, when combining these two modules, we observe a 0.7\% improvement concerning top-1 accuracy, showing the effectiveness of learning motion representation comprehensively. We also compared our CME with another two channel enhancement methods MEM and PEM with the same baseline. Owing to the global temporal receptive field when selecting discriminative channels, our CME outperforms MEM and PEM by 3.2\% and 1.9\%, respectively.

\subsection{Visualization}
We attain further insight by studying how CME and SME capture the motion information in videos.  The video frames displayed are all from the Something-Something dataset.

\begin{figure}[tb!]
    \centering
    \includegraphics[width=.42\textwidth]{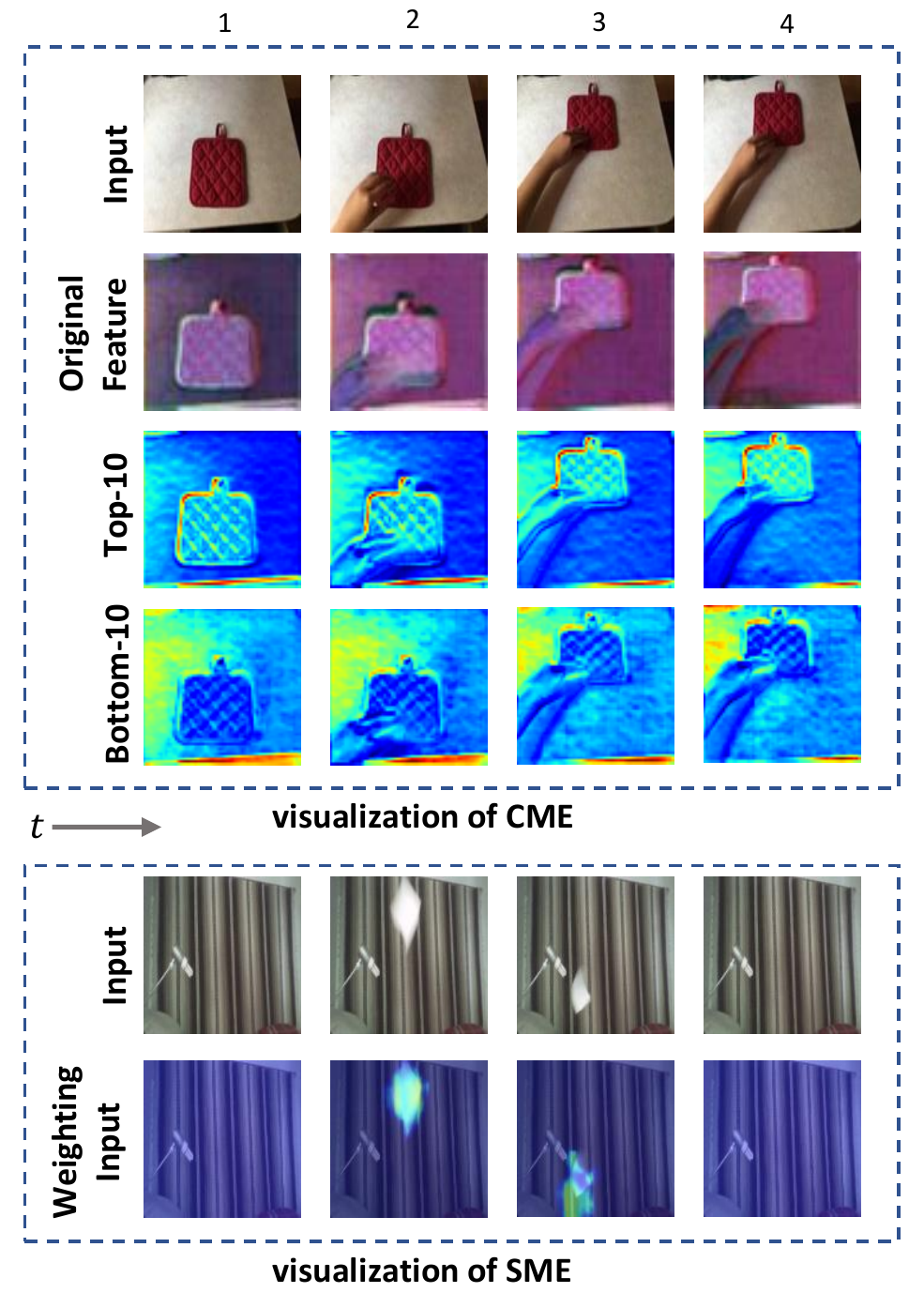}
    \caption{Visualizations of CME and SME. The action of moving an object is analyzed in the CME part: The third row shows the channels selected by CME to strengthen, and the fourth row displays the channels to suppress. A scene of paper falling is discussed in SME visualization, and the spatial-wise attention map is shown in the last row.}
    \label{fig:visual-ALL}
\end{figure}

For CME, we compare the channels enhanced by CME with the ones that are suppressed. We extract the features before the CMR-BlockB in the first stage of the base model. The channels of each frame are then sorted in descending order according to the generated channel importance weights. We take the top-10 and bottom-10 of these channels as two groups and merge each group to a single channel by an average operation. The single-channel feature is used to generate the heat map. As illustrated in the top half of Fig.~\ref{fig:visual-ALL}, we display the heat maps of the top-10 and bottom-10 groups in the third and fourth row, respectively. It is obvious that top-10 channels, which will be strengthened by the CME module, intend to focus on the moving target. In contrast, the bottom-10 channels are sensitive to the static background. It indicates that our CME module has a strong ability to distinguish motion-related channels and background-related channels.
\begin{table}[tb!]
    \centering	\resizebox{0.46\textwidth}{!}{
    \begin{tabular}{lccc}
        \toprule
         \textbf{Method}&\textbf{Top-1}&\textbf{Top-5}&\textbf{Latency}\\
         \midrule
        
        baseline$+$CME$_{all}$$+$SME$_{all}$&51.0\%&79.3\%&30.4 ms \\
         baseline$+$CME$_{all}$$+$SME$_{part}$&\textbf{51.3\%}&\textbf{79.8\%}&21.4 ms \\
         baseline$+$CME$_{part}$$+$SME$_{all}$&51.2\%&79.2\%&26.5 ms \\
         baseline$+$CME$_{part}$$+$SME$_{part}$&50.4\%&78.6\%&\textbf{17.4} ms \\
         \bottomrule
    \end{tabular}}
    \caption{Ablation study for effect of the inserted module. The baseline is ResNet-50 equipped with TIM. The “all” means this module is embedded into each block of ResNet and the “part” means only each stage’s first block is equipped with the module.}
    \label{tab:ablation_num}
\end{table}

\begin{table}[tb!]
    \centering
    \begin{tabular}{lcc}
        \toprule
         \textbf{Method}&\textbf{Top-1}&\textbf{Top-5}\\
         \midrule
         baseline~\cite{liu2020teinet}&46.1\%&74.7\%\\
         $+$MEM~\cite{liu2020teinet}&47.4\%&76.6\%\\
         $+$PEM~\cite{weng2020temporal}&48.7\%&77.8\%\\
         $+$CME&50.6\%&79.4\% \\
         $+$SME&50.6\%&78.9\% \\
         $+$CME\&SME&\textbf{51.3\%}&\textbf{79.8\%}\\
         \bottomrule
    \end{tabular}
      \caption{Ablation study: effectiveness of each module.}
    \label{tab:ablation_tim}
\end{table}

For SME, we visualize the weighting map generated by SME in the first stage of the model. As shown in the bottom half of Fig.~\ref{fig:visual-ALL}, only the moving object (paper) will be highlighted, avoiding overfitting to the background (table lamp).

\begin{table}[tb!]
    \centering
    \small
\begin{tabular}{lcccc}
\toprule 
\textbf{Model} & \textbf{L} & \textbf{T} & \textbf{A}\tabularnewline
\midrule 
TSM \cite{lin2019tsm} &  12.3 & 107.3 & 45.6\tabularnewline
TEI \cite{liu2020teinet} & 20.4 & 78.5 & 47.4\tabularnewline
STM \cite{jiang2019stm} & 14.7 & 93.2 & 49.2\tabularnewline
TDRL \cite{weng2020temporal}  & 63.8 & 49.5 & 49.8\tabularnewline
\midrule
baseline$+$CME  & 18.9 & 75.2 & 50.6\tabularnewline
baseline$+$SME  & 16.3 & 72.6 & 50.6\tabularnewline
baseline$+$CME\&SME  & 21.4 & 61.8 & 51.3\tabularnewline
\bottomrule
\end{tabular}    
\caption{Runtime comparison on Something-Something V1 dataset. L denotes Latency (ms), T denotes Throughput (videos/s), A denotes Top-1 Accuracy (\%). The baseline is ResNet-50 equipped with TIM.}
    \label{tab:runtime}
\end{table}

\subsection{Runtime Analysis}

We follow the inference settings in \cite{lin2019tsm} by using a single NVIDIA Tesla V100 GPU to measure the latency and throughput. We use a batch size of 1 and 8, to measure the latency and throughput, respectively. The results are shown in Table~\ref{tab:runtime}. Our method achieves similar latency with TEI but outperforms it by 3.9\%. Compared with the state-of-the-art method TDRL, our model is nearly three times faster and obtains 1.5\% improvement on accuracy.

\section{Conclusion and Outlook}
In this paper, we have proposed the CMR-Block for comprehensive motion representation learning, which consists of Channel-wise Motion Enhancement Module and Spatial-wise Motion Enhancement Module. Specifically, the CME is designed to enhance the discriminative channels and suppress the useless ones from the global temporal receptive field. The SME is designated to highlight the features in motion area based on the similarity map between two adjacent frames. The experiments are conducted on three benchmark datasets to demonstrate the effectiveness of our proposed method. 
On two temporal reasoning datasets, our method achieves the state-of-the-art results by a large margin with a little additional computation cost.

In future work, we are exploring the way of integrating optical flow networks into our learning pipeline, especially for those unsupervised flow models based on our previous work~\cite{ren2017unsupervised,ren2020stflow,ren2020unsupervised}, as the two tasks for action recognition and flow estimation can be of mutual benefit to each other.
\section{Acknowledgements}
This work is in part supported by National Key Research and Development Program of China (2018AAA0100704), NSFC (U19B2035, 61972250, 72061127003, U20B2068) and Shanghai Municipal Science and Technology Major Project (2021SHZDZX0102).

\section{Ethic Statement}
This paper aims to improve the precision and processing speed in video understanding technology, which is widely used in industry and our daily life. Efficient automated video processing greatly reduces the workload of human beings and also helps people to obtain valuable information more conveniently. However, as the applications of this technology penetrate into more fields, personal privacy will be under threat. Therefore we should pay attention to the privacy issues and other potential social security issues that may arise during the development of such technology.

\bibliographystyle{aaai21}
\bibliography{main}

\end{document}